# SENTIMENT CLASSIFICATION OF CODE-SWITCHED TEXT USING PRE-TRAINED MULTILINGUAL EMBEDDINGS AND SEGMENTATION

Saurav K. Aryal[1] Howard Prioleau[1] and Gloria Washington[1]

[1]Department of Electrical Engineering and Computer Science, Howard University, Washington DC, USA
saurav.aryal@howard.edu
howard.prioleau@bison.howard.edu
gloria.washington@howard.edu

## ABSTRACT

*With increasing globalization and immigration, various studies have estimated that about half of the world population is bilingual. Consequently, individuals concurrently use two or more languages or dialects in casual conversational settings. However, most research is natural language processing is focused on monolingual text. To further the work in code-switched sentiment analysis, we propose a multi-step natural language processing algorithm utilizing points of code-switching in mixed text and conduct sentiment analysis around those identified points. The proposed sentiment analysis algorithm uses semantic similarity derived from large pre-trained multilingual models with a handcrafted set of positive and negative words to determine the polarity of code-switched text. The proposed approach outperforms a comparable baseline model by 11.2% for accuracy and 11.64% for F1-score on a Spanish-English dataset. Theoretically, the proposed algorithm can be expanded for sentiment analysis of multiple languages with limited human expertise.*

## 1. INTRODUCTION

Linguistic code-switching is the concurrent use of two or more languages or dialects in a conversation. According to The International Organization for Migration's (IOM) World Immigration Report, physical migration only accounts for 3.1% of the world population in 2020. However, growth in internet technology has introduced a new wave of unfiltered and cross-cultural-lingual interactions, which provide a fertile ground for linguistic code-switching. Therefore, sentiment analysis of linguistic code-switching is a critical task in the future of multilingual natural language processing and understanding.

Code-switching, in general, has been studied extensively in psycholinguistics and sociolinguistics [1,2,3]. While the psycholinguistic definition of code-switching is multi-faceted, this paper, for readability and convenience, will refer to only the linguistic variant as code-switching henceforth.

While research in sentiment analysis has proliferated over the past decade, most research focused on utilizing monolingual textual data. Consequently, sentiment analysis of code-switched text is limited. Furthermore, sentiment analysis of code-switched data is a more complex task because two languages' unification creates a tertiary low-resource language. This low-resource language maintains the two primary languages' features while creating a new topology that classification models must learn separately.

To further the work in code-switched sentiment analysis, we propose a multi-step natural language processing algorithm utilizing points of code-switching in mixed text and conduct sentiment analysis around those identified points. The proposed sentiment analysis algorithm uses semantic similarity derived from large pre-trained multilingual models with a handcrafted set of positive and negative words to determine the polarity of code-switched text. Additionally, the proposed algorithm can be adapted for sentiment analysis of multiple languages with limited human expertise.

## 2. RELATED WORKS

In this section, we will review literature related to transformer-based architectures for sentence embeddings and sentiment analysis of code-switched text.

Extracting embeddings is the process of mapping textual data to vectors of real numbers. These embeddings are learnt representations of sentences and can be used in semantic similarity. Bidirectional Encoder Representations from Transformers (BERT) [4] and related variants [5,6] have provided state-of-the-art embeddings for numerous natural language processing (NLP) tasks. While the seminal BERT does not compute sentence-level embeddings, researchers have trained BERT model for sentences [7,8,9]. However, BERT models are computationally expensive for semantic similarity searches, Siamese-BERT (SBERT) [10] enabled efficient comparisons. Various multi-lingual SBERTs have been applied code-switched text [11]. The findings of [11] suggest that a pre-trained multi-lingual model does not necessitate high-quality representations of code-switched text.

In another line of work, bilingual embeddings have been introduced to represent code-switching sentences [12,13,14,15,16]. However, these approaches are limited to fewer languages. Independently, much work has been done on extracting sentence embeddings [17,18,19,20]. Of these, Universal Sentence Encoder [20] was trained with the explicit goal of learning multilingual embeddings for downstream tasks.

Despite available embedding methods and accessible code-switched social media text, few labelled data sets have been released for sentiment analysis. These datasets include Malayalam-English and Tamil-English [21,22,23], Spanish-English [24], and Hindi-English [25].

Previous work has shown BERT-based models achieve state-of-the-art performance for code-switched languages in tasks like offensive language identification [26] and sentiment analysis [27]. Custom architectures and algorithms have also been proposed [24,28,29].

Of these approaches we chose [24] as our baseline of comparison since this model, to the best of our knowledge, is the current state-of-art for this specific dataset-task pair and has shown improved performance when compared to over 10 other approaches. Furthermore, other models have only explicitly focused on Dravidian languages which is likely easier than Spanish-English sentiment analysis due to difference in lexical relatedness [30,31].

In our work, we explore the possibility of exploiting multi-lingual embedding spaces toward sentiment analysis of code-switched text. Our proposed algorithm can be extended in a language-agnostic manner with nominal human involvement.

## 3. METHODOLOGY

### 3.1. Datasets

To analyze this project, we utilize two primary datasets. Firstly, the Spanish-English Code-switching Dataset [24] sourced from tweets that contained both Spanish and English. Secondly, the Word Level Sentiment Lexicon [32] dataset which provides a lexicon of negative and positive words for many languages. For this work, we only utilize the Spanish and English subsets.

### 3.2. Data Preprocessing

We convert the text data to lower case and remove, to the best of our abilities, all punctuation, white spaces, and sensitive information such as email addresses, URLs, numbers, and social media entities like emojis and hashtags.

### 3.3. Multilingual Models

In this work, we compare two standard, pre-trained, multilingual model for extracting sentence-level embeddings: SBERT-based model XLM-Roberta (XLM-RBASE) [10] and Universal Sentence Encoder (USE) developed by Google in [20]. Our algorithm's intuition relies on the hypothesis that sentence embeddings of segments that are negative sentence will be closer in vector space to a set of negative ground truth words/phrases and vice versa for positive sentences and their corresponding ground words despite their language of origin.

### 3.4. Code-switching Identification Algorithm

The proposed system in this work is composed of two primary parts. The first part is the detection of code-switch points in text. To implement this, we utilized a Naïve Bayes classifier [33] on our Spanish-English dataset to detect likely points where code-switching might occur. In their dissertation research, [33] seeks to detect code-switching points in several interviews where the interviewer and the interviewee switch between English and Kiswahili. For reference, an example of code-switching in Swahili-English points can be found in Table 1.

| Original Text | Translation | Code-Switching Points |
|---|---|---|
| Okay, Okay, na unafikiria ni important kujua native language? | Okay, and do you think it is important to know native language? | "okay," "ni," "important," and "kujua" |

Table 1: Code-switching points represent point at which the language of the sentence switches from one language to another. This table show examples of code-switching points in various code-mixed language combinations.

```
pos_eng: The list of ground positive English words
neg_eng: The list of ground negative English words
pos_es: The list of ground positive Spanish words
neg_es: The list of ground negative Spanish words

# compare the language-segments with negative and positive word sets
scores = cosineSimilarity(spanish_segments, pos_es)
neg_scores = cosineSimilarity(spanish_segments, neg_es)
eng_scores = cosineSimilarity(english_segments, pos_eng)
eng_neg_scores = cosineSimilarity(english_segments, neg_eng)

# get the total Positive and Negative score
neg = sum(neg_scores, eng_neg_scores)
pos = sum(scores, eng_scores)

# Thresholding
if pos >neg:
   return "Positive"
elif neg <pos:
   return "Negative"
```

Table 2: Pseudo Code for English-Spanish Text

### 3.5. Sentiment Analysis Algorithm

The code-switch points are passed into the second part of the overall system for sentiment analysis. Using these results, the code-switched text is segmented into constituent languages. Assuming the two languages involved are know, which for our dataset is English and Spanish, the segments embeddings are extracted independently using USE and XLM-RBASE. A set of positive and negative sentiment lexicons [32] for each language and their embeddings are also extracted. Segments were compared to their language-specific sets of negative and positive embeddings. The returned comparison from the four sets of comparisons from two languages each with two sets of curated embeddings is the totalled and the total negative score and total positive scores are used to determine the polarity of the overall codeswitched sentence. The pseudocode algorithm for detecting the sentiment post-segmentation into English and Spanish can be seen in Table 2. For clarity, we summarize the full proposed algorithm in Figure 1 below.

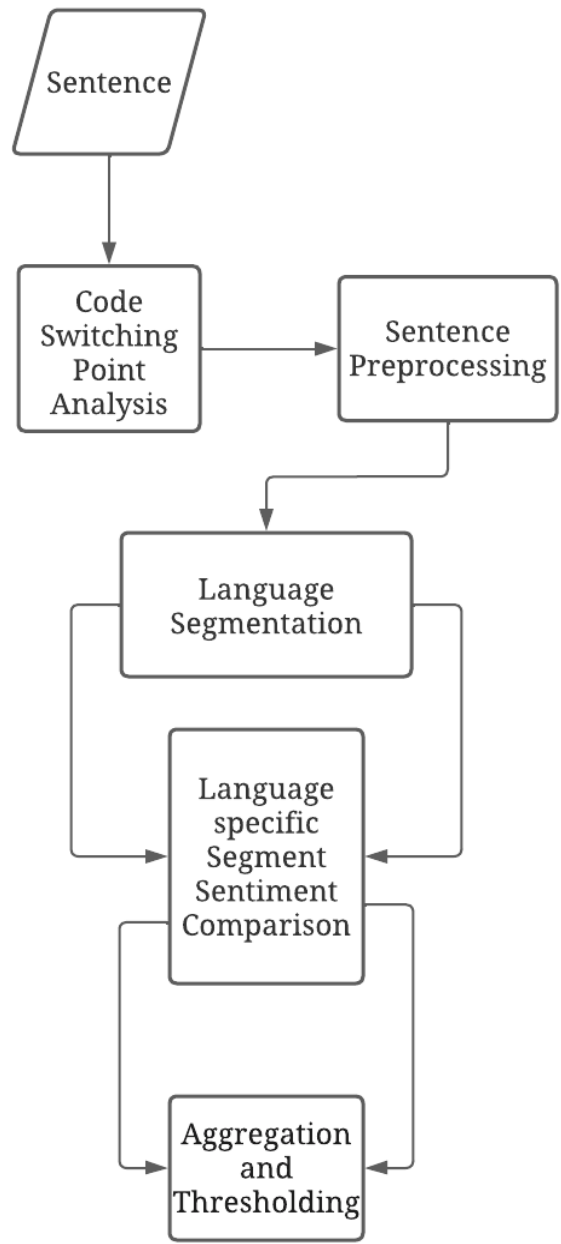

Figure 1: Full Proposed Algorithm

To enable future comparisons and ensure reproducibility, a link to access our source code has been provided in the *Appendix*.

## 4. EXPERIMENTAL RESULTS

| Model | Accuracy | Macro F1 | Precision | Recall |
|---|---|---|---|---|
| GIRNET 1L | 63.3 | 62.4 | 63.4 | 61.8 |
| GIRNET 2L | 62.0 | 61.3 | 61.8 | 61.1 |
| USE | 71.9 | 71.7 | **74.5** | 73.4 |
| XLM-RBASE | **74.5** | **74.0** | 74.2 | **73.8** |

Table 3: Comparison of results

### 4.1. Sentiment Analysis Evaluation

Table 3 shows the sentiment analysis results via the semantic similarity task. This table compares our results with other methods for classifying the sentiment of the code-switched text. The first two rows show the results provided by the baseline [24]. The baseline results from the paper, which utilize a unified position-sensitive multi-task recurrent neural network (RNN), produce an accuracy of 63.3% and a macro-F1 score of 62.36% on the Spanish-English code-switched dataset. For context, in table 3, GIRNET 1L represents the architecture with one layer, whereas GIRNET 2L illustrates the architecture with two layers. The third and fourth columns show results from our algorithm when utilizing Universal Sentence Encoders (USE) and XLM-RBASE. For further analysis, we have provided the confusion matrix of the approach utilizing XLM-RBASE and USE in Tables 4 and 5 respectively. In terms of overall performance, we notice that XLM-RBASE slightly outperforms USE embeddings. However, USE embeddings are better for predicting positive sentiments than XLM-RBASE embeddings but much worse for negative sentiments. Furthermore, false positive and false negative predictions of the best-performing model have been provided the *Appendix* for further qualitative analysis.

| | True Positive | True Negative |
|---|---|---|
| Predicted Positive | 429 | 201 |
| Predicted Negative | 160 | **624** |

Table 4: XLM-RBASE Model Confusion matrix

| | True Positive | True Negative |
|---|---|---|
| Predicted Positive | **553** | 77 |
| Predicted Negative | 321 | 463 |

Table 5: USE Model Confusion matrix

### 4.2. Future Work and Limitations

While the proposed approach significantly improves the baseline's performance metrics for the task, the model's performance still leaves more to be desired. Furthermore, while our proposed approach works on English-Spanish, the work done explicitly on Dravidian languages or utilizing domain-specific dependence are not directly comparable. Additionally, even though these languages are prominently studied in code-switching literature, all current datasets only contain code-switching between English and one other language whereas other relevant language pairs such as French-Tamil, Nepali-English, Hopi-Tewa, Swahili-English, Hindi-Nepali, and Latin-Irish currently lack datasets. Thus, much work needs to be done in creating datasets that go beyond the language pairs currently being studied.

The algorithm proposed still depends on models that can provide accurate representations across the languages considered and assume the languages involved have already been identified. This requirement can be a particular issue for other low-resource languages. Furthermore, while the given approach requires limited human intervention, we may still need human expertise to create positive and negative word sets for the language of choice. We hope that with increasing need and interest for this topic, larger and more challenging datasets and increased research funding will follow in the future.

Finally, the authors would like to acknowledge that the approach is still rather rudimentary and should definitely be improved. However, given that this seemingly simple approach outperformed the extensively tested and task-specific custom model of the baseline, perhaps future research in this topic needs to take a new outlook beyond novel and custom neural network architectures while recognizing the niche task at hand.

## 5. CONCLUSION

In this work, we present our research towards developing a pipeline to detect sentences where code-switching present and to classify the polarity of these sentences. This pipeline has the potential to assist researchers in further improving the sentiment analysis of code-switched text.

The results show that our approach significantly surpasses the baseline performance (GIRNET) by 11.2% for accuracy and 11.6% for the F1 -score. Furthermore, when comparing the two embedding models utilized within our algorithm, XLM-RBASE outperformed the Universal Sentence Encoder. These results suggest that our algorithm utilizing semantic similarity and multilingual language models for representation can offer a method for detecting the sentiment of code-switched text in a language agnostic manner. Theoretically, the proposed approach may be applied to a number of downstream classification tasks and applications that involve code-switched textual data. However, larger and comprehensive data sets may enable more complex and better performing approaches for language-agnostic sentiment analysis of code-switched text. To be added if accepted.

## ACKNOWLEDGEMENTS

THIS PROJECT WAS SUPPORTED (IN PART) BY FROM THE NORTHROP GRUMMAN CORPORATION UNDER AWARD #201603 AND THE NATIONAL SECURITY AGENCY, USA UNDER AWARD NUMBER H98230-19-P-1674. THE AUTHORS WOULD ALSO LIKE TO ACKNOWLEDGE THE INITIAL WORK AND SUPPORT PROVIDED BY MR. CESA SALAAM IN THE PREPARATION OF THE MANUSCRIPT. THE CONTENT IS SOLELY THE RESPONSIBILITY OF THE AUTHORS AND DOES NOT NECESSARILY REPRESENT THE OFFICIAL VIEWS OF THE NORTHROP GRUMMAN CORPORATION, OR THE NATIONAL SECURITY AGENCY.

for code-switched data', *arXiv preprint arXiv:2102. 12407*, 2021.

## APPENDIX

All supplemental materials (source code, data set, results) can be found in the link below:

https://bit.ly/3fZrMNJ